\documentclass[sigconf,authorversion,nonacm]{acmart}

\copyrightyear{2024}
\acmYear{2024}
\setcopyright{acmlicensed}\acmConference[HRI '24]{Proceedings of the 2024 ACM/IEEE International Conference on Human-Robot Interaction}{March 11--14, 2024}{Boulder, CO, USA}
\acmBooktitle{Proceedings of the 2024 ACM/IEEE International Conference on Human-Robot Interaction (HRI '24), March 11--14, 2024, Boulder, CO, USA}
\acmPrice{15.00}
\acmDOI{10.1145/3610977.3634933}
\acmISBN{979-8-4007-0322-5/24/03}

\usepackage{subfig}
\usepackage{graphicx}

\begin{document}

\title{Social Cue Detection and Analysis Using Transfer Entropy}

\author{Haoyang Jiang}
\affiliation{%
  \institution{Monash University}
  \streetaddress{Wellington Rd, Clayton}
  \city{Melbourne}
  \state{Victoria}
  \country{Australia}
  \postcode{3800}
}
\email{haoyang.jiang@monash.edu}

\author{Elizabeth A. Croft}
\affiliation{%
  \institution{University of Victoria}
  \streetaddress{3800 Finnerty Rd}
  \city{Victoria}
  \state{British Columbia}
  \country{Canada}
  \institution{Monash University}
  \streetaddress{Wellington Rd, Clayton}
  \city{Melbourne}
  \state{Victoria}
  \country{Australia}
}
\email{ecroft@uvic.ca}

\author{Michael G. Burke}
\affiliation{%
  \institution{Monash University}
  \streetaddress{Wellington Rd, Clayton}
  \city{Melbourne}
  \state{Victoria}
  \country{Australia}
  \institution{University of Edinburgh}
  \streetaddress{Old College, South Bridge}
  \city{Edinburgh}
  \state{Scotland}
  \country{United Kingdom}
}
\email{michael.g.burke@monash.edu}

\renewcommand{\shortauthors}{Haoyang Jiang, Elizabeth A. Croft, \& Michael G. Burke}

\begin{abstract}
Robots that work close to humans need to understand and use social cues to act in a socially acceptable manner. Social cues are a form of communication (i.e., information flow) between people. In this paper\footnote{This paper has been accepted by HRI'24.}, a framework is introduced to detect and analyse a class of perceptible social cues that are nonverbal and episodic, and the related information transfer using an information-theoretic measure, namely, transfer entropy. We use a group-joining setting to demonstrate the practicality of transfer entropy for analysing communications between humans. Then we demonstrate the framework in two settings involving social interactions between humans: object-handover and person-following. Our results show that transfer entropy can identify information flows between agents and when and where they occur. Potential applications of the framework include information flow or social cue analysis for interactive robot design and socially-aware robot planning.
\end{abstract}




\maketitle

\section{Introduction}\label{intro}
Social robots are a category of robots that work physically close to humans and are designed to interact with non-expert users. User acceptance requires that these robots operate in a socially acceptable manner and are able to react to, or exchange, social cues that play an important role during interaction. While people frequently exchange social cues using verbal and gestural signals, they also exchange rich information through their gait, posture and walking patterns \cite{rios-martinez_proxemics_2015}. For instance, in a handover task, a giver typically reaches forward to indicate their intent to give an item and a receiver observes this cue and reaches forward indicating readiness to accept the item. Many nonverbal, subtle cues like this reaching cue might not be immediately apparent to a casual observer, but play an important role when people collaborate or interact with each other \cite{vinciarelli_social_2008}.

Socially-aware behaviour is a multi-lateral process, and to be useful in robots it needs to be predictable, adaptable and easily understood by humans \cite{rios-martinez_proxemics_2015}. This requires that social robots sense and react to social information conveyed by humans while simultaneously conveying useful social information to humans. However, due to hardware limitations, environmental constraints and the robot's primary task, not all robots can communicate through dedicated audio or visual display channels. Therefore, being able to capture and use measurable social cues embedded in kinematic measurements (e.g., pose and motion) is an important step for social robots to achieve more acceptable and anthropomorphic behaviours while working with humans. Detecting and reacting to subtle social cues is crucial to move beyond explicit and often contrived or overt human-robot interaction. Unfortunately, this is very challenging because humans or agents can express themselves in a variety of ways, which means that the cues themselves are often difficult to elucidate. This work seeks to address this challenge by defining a class of measurable social cues and proposing a framework that allows these cues to be detected automatically.

In signal detection theory, Green and Swets \cite{green_signal_1966} introduced the concept of \textit{response bias}, which refers to how much evidence an observer requires before responding to a given signal. Our cue detection framework builds on this theoretical notion. In our framework, the given signal refers to social cues. If a cue is below the response bias of the observer, it is a silent cue, meaning that one cannot tell if a cue exists by solely observing the observer. For the purposes of this work, we focus on those cues that are above the response bias. We define a \textit{perceptible social cue} as \textit{an event generated by an agent that influences the behaviour of the other agent observing the cue}. This paper aims to formulate a method to detect perceptible cues of this form.

The proposed perceptible social cue analysis framework accepts raw information captured by sensors on robots, and seeks to primarily provide answers to the following questions:
\begin{itemize}
    \item When or where is a perceptible social cue activated?
    \item What is the direction of the exchanged cue?
\end{itemize}
A core contribution of the proposed framework is to model implicit communication between agents, and thereby the exchange of perceptible social cues, as an \textit{exchange of information}. This allows us to use information-theoretic approaches to measure continuous information flows between agents, and threshold these to identify perceptible social cues. Herein, cue transfer is analysed using an information-theoretic measure, Transfer Entropy (TE), a statistical measure of the amount of directed transfer of information between two systems \cite{schreiber_measuring_2000}. Other contributions include:
\begin{itemize}
    \item a general framework that is able to detect arbitrary cues from raw data of social interactions without pre-designing or predefining a set of cues.
    \item a method of computing TE locally with multi-dimensional features using neural networks.
\end{itemize}
We validate the framework in three unique settings: group joining, handover and person-following, to showcase the broad applicability of perceptible social cue analysis.

\section{Background}
\subsection{Social Cues} 
It is broadly recognised that social cues play an important role in communication. However, for social cue detection, researchers usually predefine a fixed set of social cues composed of known postures and gestures such as nodding or head shaking \cite{urakami_nonverbal_2023}\cite{bousmalis_towards_2013}\cite{bremers_using_2023}. NovA \cite{hutchison_nova_2013} is a well-developed system for nonverbal signal detection, which detects postures and gestures using event-based gesture analysis \cite{kistler_natural_2012}, and classifies them into a predefined set of cues. This system also provides a movement expressivity measurement. However, these measurements are not used for cue detection.

Social cues are widely used in robotics, especially for Human-Robot Interaction (HRI). Tomari et al. proposed a socially-aware navigation planner for wheelchair robots, tracking the head orientations of the participants to estimate personal space assuming humans are more protective of the space in front of them \cite{tomari_analysis_2014}. Hansen et al. developed an adaptive system for natural interaction between mobile robots and humans based on the person's pose and position, which estimates the interaction intention of the user and then uses it as a basis for socially-aware navigation based on a person's social space \cite{tranberg_hansen_adaptive_2009}. Escobedo et al. used the commonly visited destinations of a wheelchair robot's user to estimate the probable intended destination of the user and then accept the user's face and voice commands for navigation \cite{escobedo_using_2014}. These works are dependent on the proxemics theory proposed by Hall, who defined general social zones of humans \cite{hall_hidden_1966}. However, the concept of social zones relies on averaged heuristics. Hall validated his study only for US citizens \cite{rios-martinez_proxemics_2015}, which means these metrics may not translate well to other cultural contexts. Many human-following robot designs use social cues, e.g., in \cite{mi_system_2016} where body orientation is used to estimate the intended turning direction. In \cite{moustris_intention-based_2016}, the relative position between a human and a robot is used as a feature to anticipate the human. Hu et al. \cite{hu_design_2014} use human orientation as an input for anticipatory robot behaviours. In general, most robots are designed to act on heuristic or manually selected social cues, and it is unclear whether these social cues are reliable or repeatable. Consequently, most current research focuses on detecting these cues to predict intent or behavior.

Cue detection usually requires tuning and is not necessarily repeatable or cross-cultural. There are no examples of research where interaction cues are automatically sourced. We tackle these issues in this work. The perceptive social cue detection framework proposed here identifies social cues both spatially and in time from raw data, requiring no predefined cue sets and comparatively less manual specification. Since it is data-driven, the framework is agnostic to social norms or heuristics and can be applied to any context where motion data can be tracked.

\subsection{Communicating intent}
The study of social cues is related to intent communication, which refers to behaviours that allow an observer to quickly and correctly infer the intention of the agent generating the behaviour \cite{busch_learning_2017}\cite{dragan_legibility_2013}. The research in \cite{lichtenthaler_influence_2012} shows that motions that communicate intent increase perceived safety during virtual human-robot path-crossing tasks. In \cite{busch_learning_2017}, the authors consider robustness, efficiency and energy as universal costs in a reinforcement learning scheme, with results showing an increase in the ability of a human to interpret a robot's intention. Research studying projected visual legibility cues \cite{hetherington_hey_2021}, has shown that projected arrows are generally more interpretable than flashing lights in a navigation setting. Importantly, the ways people communicate and understand intent are different, and there is no universal method to measure the communication of intent. However, if communicating intent is interpreted as the transfer of information between agents, this concept can be captured by the proposed framework, which could be used as a standard for measurement and to guide the selection of robot motions or actions.

\subsection{Transfer Entropy} 
TE is a measure that allows the analysis of the information transfer and potential causal relationships between two simultaneous time series. In the economics literature, Baek et al. use TE to analyse the market influence of companies in the U.S. stock market \cite{baek_transfer_2005}. He and Shang \cite{he_comparison_2017} compare different TE methods for analysing the relationship between 9 stock indices from the U.S., Europe and China. TE has been used to analyse animal-animal or animal-robot interactions such as \cite{shaffer_transfer_2020}\cite{porfiri_inferring_2018}. It has also been adopted to analyse joint attention \cite{sumioka_causality_2007} and model pedestrian evacuation \cite{xie_detecting_2022}. Berger et al. apply TE in robotics to detect human-to-robot perturbations using low-cost sensors \cite{berger_transfer_2014}. The above studies succeed in quantizing the information transfer or the relationship between their targets using TE, but tend to focus on specific features or aspects of interest. Our work seeks to provide a more general framework for detecting and analysing perceptible social cues, by looking for changes in information transfer over time and in space. 

Mutual information (MI) is an information measure that captures the shared information between random variables. Klyubin et al. proposed the concept of empowerment for intrinsic motivation for reinforcement learning, which measures the maximum MI (channel capacity) between the agent's actuation and their sensors \cite{klyubin_empowerment_2005}. This is expanded by Mohamed and Rezende with a lower complexity maximisation approach to MI \cite{mohamed_variational_2015}. Jaques et al. use MI as a social influence reward for multi-agent deep reinforcement learning in Sequential Social Dilemmas (SSDs) \cite{leibo_multi-agent_2017} to encourage collaborations between agents \cite{jaques_social_2019}. While MI measures the shared information, TE measures the time-asymmetric information transfer. Unlike MI, the asymmetric property of TE allows us to analyse the directionality of information flow, which is beneficial for analysing the exchange of social cues. By applying TE, we aim to quantize social cue transfer to better understand and use social cues for robots. In addition, most TE-related research reduces multi-dimensional features down to a single dimension to compute TE, while in practice, most features are multi-dimensional. We tackle this issue by using neural networks to build the probability distributions needed to estimate TE.

\section{Methodology}

\begin{figure*}[ht!]
\centerline{\includegraphics[trim={0.2cm 2.5cm 0.3cm 2.3cm},clip,width=\linewidth]{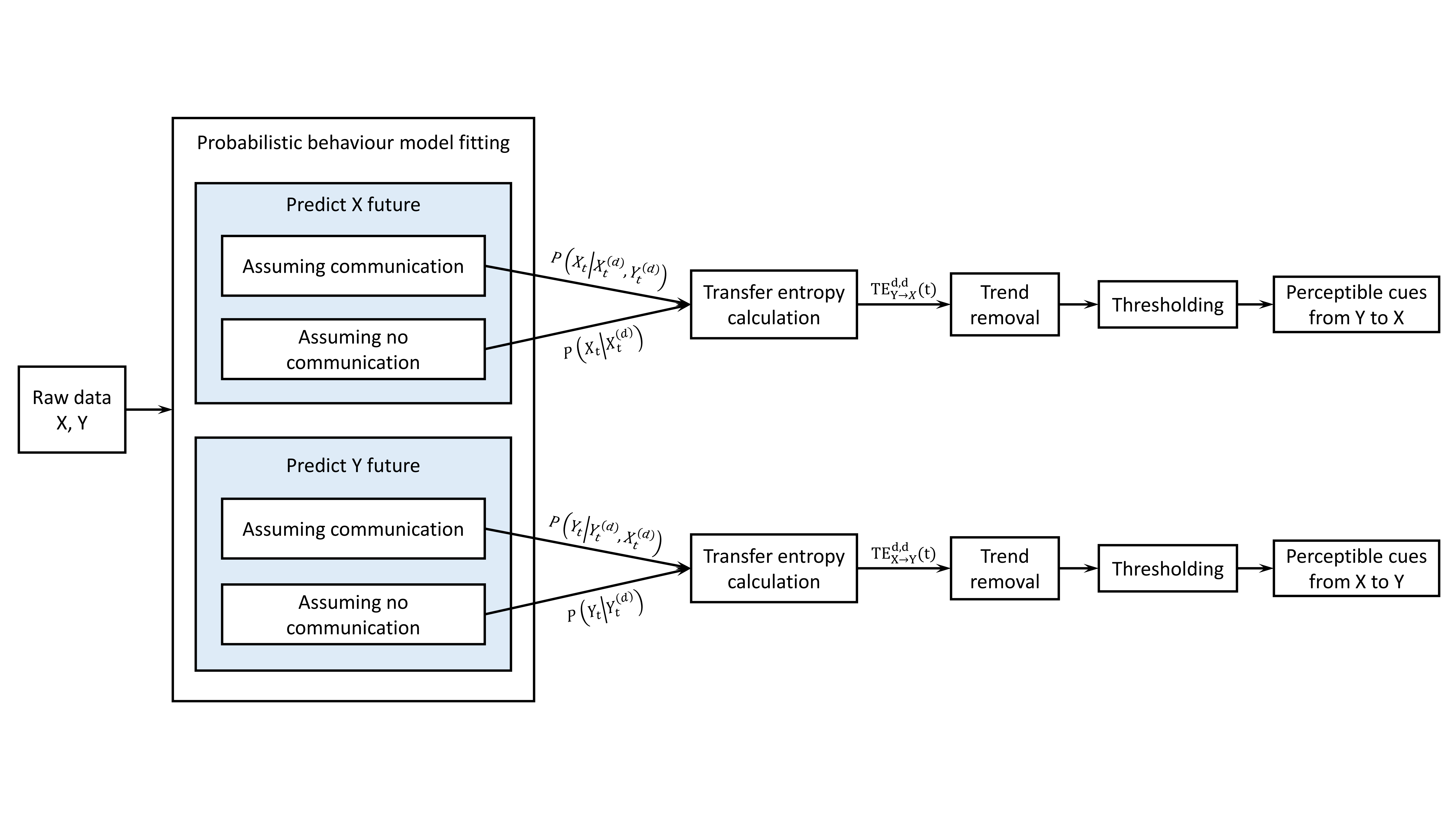}}
\setlength\abovecaptionskip{-27pt}
\caption{Illustration of the proposed framework. $X$, $Y$ represent two agents in a social interaction scenario. Raw data is used to fit probabilistic behaviour model with different conditions, the output conditional probabilities are then used to compute transfer entropy (TE) in both information transfer directions. Double exponential smoothing (DES) is used to remove the general TE trend followed by the thresholding in order to find perceptible cues using the TE.}
\label{framework}
\Description{A flow diagram illustration of the proposed framework. Raw data of two agents $X$, $Y$ is used to fit probabilistic behaviour model with conditions assuming and not assuming communication to predict the future of $X$, $Y$, and the output conditional probabilities of $X$, $Y$ future behaviour are then used to compute transfer entropy (TE) in both information transfer directions. Double exponential smoothing (DES) is used to remove the general TE trend followed by a thresholding step in order to find perceptible cues using the TE for both information transfer directions.}
\end{figure*}

We model a perceptible cue as an exchange of information between two agents, using information-theoretic approaches to measure the levels of information exchanged. While it is difficult to measure extremely subtle information transfer, given sensor capability limits and the varied time horizons over which cues can occur, it is possible to identify a restricted subset of cues within the information stream. To enable this, we make the following assumptions:
\begin{itemize}
    \item To be detected, a perceptible cue needs to exceed some significance level or base threshold of information transfer between agents.
    \item To be detected, a perceptible cue needs to occur within a finite time period.
\end{itemize}
Our framework is illustrated in Fig. \ref{framework} and described below.

\subsection{Transfer Entropy}
Transfer entropy is closely related to the concept of \textit{Wiener-Granger causality}\footnote{These are equivalent when linear Gaussian models are used.} \cite{granger_investigating_1969}. TE, $T_{Y\rightarrow X}$, can be defined as the conditional mutual information between two variables $X_t$ and $Y_t$ \cite{bossomaier_transfer_2016}, which is formulated as follows.
\begin{equation}
    \label{TE}
    \begin{aligned}
        T_{Y\rightarrow X}^{(k,l)}(t) & =I(X_t:\mathbf{Y}_{t}^{(l)}|\mathbf{X}_{t}^{(k)}) \\
        & =H(X_t|\mathbf{X}_{t}^{(k)})-H(X_t|\mathbf{X}_{t}^{(k)},\mathbf{Y}_{t}^{(l)})
    \end{aligned}
\end{equation}
Here, $I$ denotes the mutual information and $H$ Shannon's entropy \cite{shannon_mathematical_1948}. This equation defines the information transfer from $Y$ to $X$, where $t$ is the time, and $k$, $l$ are the history length of $X_t$ and $Y_t$. In some literature, $X$ is termed the target and $Y$ the source. TE can be intuitively interpreted as the reduction of uncertainty in a state $X$ predicted solely based on its own history when an additional information source $Y$ is introduced \cite{bossomaier_transfer_2016}. The key idea of the proposed framework is to use TE to measure the information transfer of perceptible social cues.

\subsection{Overall framework workflow}\label{workflow}
\underline{\textbf{1. Preparation.}}
We start by identifying a target $X$ and source $Y$. In social cue analysis scenarios, the source can be a feature that transmits cues of interest (e.g., a pedestrian's head orientation), and the target features that would be influenced by the cues (e.g., another pedestrian's pose). Next, time series data is collected for both the target and source. Prior to modelling, we use Takens' delay embedding \cite{takens_detecting_1981} to create higher-dimensional embeddings for the time-history series of features,
\begin{equation}
    \label{embedding_example}
    \begin{aligned}
        & \mathbf{X}_{t}^{(d)} = (x(t-\delta), x(t-2\delta), ..., x(t-(d-1)\delta))\\
        & \mathbf{Y}_{t}^{(d)} = (y(t-\delta), y(t-2\delta), ..., y(t-(d-1)\delta)) ,
    \end{aligned}
\end{equation}
where $d$ is the history length and $\delta$ is the unit time step. The history length for the target and source do not necessarily need to be the same, one can also denote them separately such as $k, l$ in Eq. (\ref{TE}). In general, the history needs to be long enough to capture the temporal relationship between cue cause and effect. In our experiments, we match this with estimates of human response times.

\underline{\textbf{2. Model the Target.}}
We next establish a baseline model, conditioned on only the history features of the target. The framework allows the application of different probabilistic prediction models. For instance, we can use a simple linear vector autoregressive model (VAR) to model the target $\mathbf{X}_{t}$. 
\begin{equation}
    \label{baseline_var}
    \begin{aligned}
        \mathbf{X}_{t}=c+\alpha \mathbf{X}_{t}^{(d)}+\epsilon
    \end{aligned}
\end{equation}
where $c$ is the constant intercept of the model, $\alpha$ is the time-invariant matrix that matches the dimension of $\mathbf{X}_{t}^{(d)}$ and $\epsilon$ an error term. We can also model the target using other modelling techniques such as a multilayer perceptron (MLP), a Gaussian process or potentially more complex neural network architectures. Regardless of the underlying model, we intentionally form the target output distribution as a conditional multivariate Gaussian with mean $f(\mathbf{X}_{t}^{(d)})$ and covariance $f^\mathbf{\sigma}(\mathbf{X}_{t}^{(d)})$,
\begin{equation}
    \label{baseline_guassian}
    \begin{aligned}
        P(\mathbf{X}_{t}|\mathbf{X}_{t}^{(d)}) \sim \mathcal{N}_{\mathbf{X}_{t}}(f(\mathbf{X}_{t}^{(d)}),f^\mathbf{\sigma}(\mathbf{X}_{t}^{(d)})),
    \end{aligned}
\end{equation}
This simplifies later TE calculations. After obtaining the base model conditioned only on the target history, the source is included to build a second model of the target. For example, assuming a vector autoregressive model, we obtain
\begin{equation}
    \label{cues_var}
    \begin{aligned}
        & \mathbf{X}_{t}=c+\alpha \mathbf{X}_{t}^{(d)}+\beta \mathbf{Y}_{t}^{(d)}+\epsilon \\
        & P(\mathbf{X}_{t}|\mathbf{X}_{t}^{(d)},\mathbf{Y}_{t}^{(d)}) \sim \mathcal{N}_{\mathbf{X}_{t}}(f(\mathbf{X}_{t}^{(d)},\mathbf{Y}_{t}^{(d)}),f^\mathbf{\sigma}(\mathbf{X}_{t}^{(d)},\mathbf{Y}_{t}^{(d)})).
    \end{aligned}
\end{equation}
Here, we illustrate the process using a vector autoregressive model, but models that capture non-linear effects could also be used. We use neural networks to model behaviours for the experiments below.

\underline{\textbf{3. Compute and Analyse Transfer Entropy.}} Modelling targets using conditional multivariate Gaussian random variables allows Shannon's differential entropy to be calculated as follows,
\begin{equation}
    \label{diff_entropy}
    \begin{aligned}
        H(\mathbf{x}) & =-\int p(\mathbf{x})\log p(\mathbf{x}) \,d\mathbf{x} \\
        & =\frac{D}{2}(1+\log (2\pi)) + \frac{1}{2}\log |\mathbf{\sigma}|
    \end{aligned}
\end{equation}
Here, $D$ denotes the dimension of the target variable. Using \eqref{diff_entropy} to calculate the entropy for each model, one can calculate the TE from the source to target as
\begin{equation}
    \label{transfer_entropy}
    \begin{aligned}
        T_{Y\rightarrow X}^{(d,d)}(t) & =I(\mathbf{X}_{t}:\mathbf{Y}_{t}^{(d)}|\mathbf{X}_{t}^{(d)}) \\
        & =H(X_t|\mathbf{X}_{t}^{(d)})-H(X_t|\mathbf{X}_{t}^{(d)},\mathbf{Y}_{t}^{(d)})
    \end{aligned}
\end{equation}

Finally, we can analyse the information transfer from the source to the target using the measured TE. 

\subsection{Computing Transfer Entropy}
The signals received from agents are usually in continuous form. Researchers often discretise these continuous signals in order to statistically compute TE using histograms or other frequentist approaches \cite{berger_transfer_2014}\cite{orange_transfer_2015}. Kernel density estimation has also been recommended in \cite{schreiber_measuring_2000} and frequently used in the information analysis toolbox \cite{lizier_jidt_2014}, a popular tool for modelling TE in economics. However, these methods require a large amount of data to build sufficiently dense probability distributions, especially if the feature dimension is high \cite{orange_transfer_2015}. Therefore, they are not helpful for local time series analysis with a limited number of samples, which means these methods are not suitable for computing local TE in order to analyse information transfer spatially or temporally. Even if we have sufficient data, building high-dimensional density distributions is not computationally efficient, which means these approaches are unsuitable for online algorithms. To tackle this issue, we use neural networks in our framework to estimate distributions, and then calculate entropy directly from the estimated conditional probabilities. This allows us to estimate TE locally and continuously, which is also computationally more efficient and achieves spatial and temporal information transfer analysis. It should be noted that since our data is continuous and entropy is computed using Shannon's differential entropy, the transfer entropy can be negative indicating the absence of information being communicated. 

\subsection{Cue detection}

\begin{figure}[ht!]
\centerline{\includegraphics[width=0.85\linewidth]{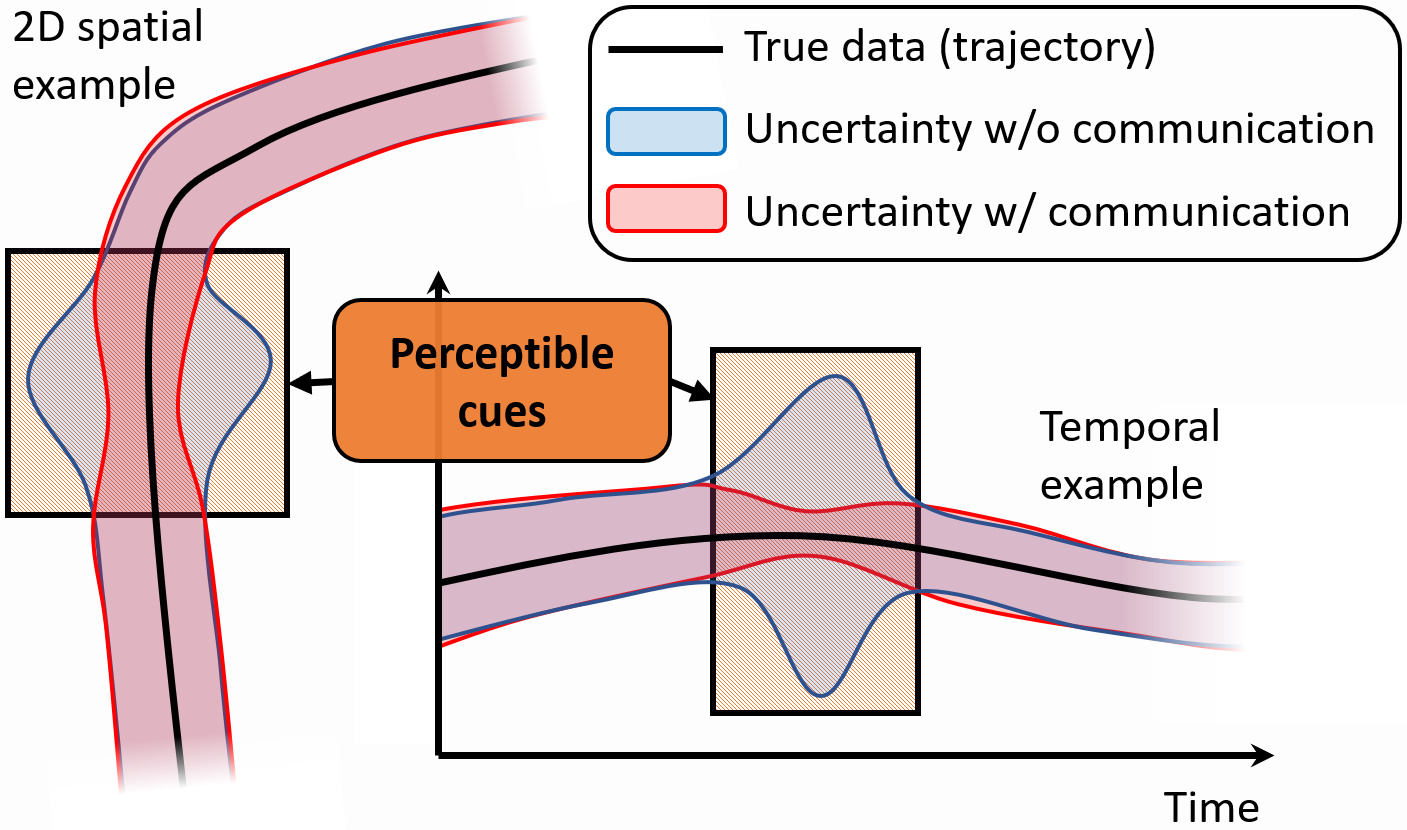}}
\setlength\belowcaptionskip{-10pt}
\caption{An illustration of the perceptible cue detection framework.}
\label{Illustration}
\Description{An illustration of the perceptible cue detection framework. Prediction uncertainties with or without communication are drawn along the true data (trajectory) for a 2D spatial example and a temporal example. When the uncertainty without communication is conspicuously larger than the uncertainty with communication, a perceptible cue is detected.}
\end{figure}

A graphical illustration of the intuition underlying perceptible cue detection is given in Fig. \ref{Illustration}. We use the TE to detect perceptible social cues.\footnote{We provide an example at \href{https://github.com/jhy9968/TEscd.git}{https://github.com/jhy9968/TEscd.git}} Positive TE means the information transfers from the source to the target. However, there can be continuous information transfer during social interactions. We are most interested in the set of special events that trigger observer responses. Therefore, to meet the criteria of a perceptible social cue in our framework, we require TE measures to meet the requirements that: 1) the TE has to be larger than zero indicating general information transfer; and 2) the TE has to be larger than a threshold to be identified as a perceptible cue. To determine the threshold, we use double exponential smoothing (DES) \cite{gardner_jr_exponential_1985} to find the moving mean and standard deviation of the TE. Let us denote $\mu_t$, $\sigma_t$ and $T_t$ as the moving mean, moving standard deviation of TE and current TE at time $t$. The DES works as follows:
\begin{equation}
    \label{DES}
    \begin{aligned}
        &\mu_t =\alpha T_t+(1+\alpha)(\mu_{t-1}+b_{t-1}) \\
        &\sigma_t = (1-\alpha)(\sigma_{t-1}+\alpha(T_t-\mu_{t-1}-b_{t-1})(T_t-\mu_{t-1})) \\
        &b_t =\beta(T_t-x_{t-1})+(1-\beta)b_{t-1}
    \end{aligned}
\end{equation}
Here, $b_t$ is the smoothed trend at time $t$, $\alpha$ is a smoothing factor and $\beta$ a trend smoothing factor. Then the perceptible cue threshold is computed as
\begin{equation}
    \label{threshold}
    \begin{aligned}
        threshold_{t} = \mu_{t-1} + \gamma\sigma_{t-1},
    \end{aligned}
\end{equation}
where $\gamma$ is a tunable parameter. We use $\gamma=3$ in the work. The thresholding process filters out TE trends, which represent the accumulated information during continuous information exchanges. TE regions that meets the requirement $TE_t > 0$ and $TE_t > threshold_t$ indicate perceptible social cues. In order to further remove the effect of the trend, we apply a 1st order high-pass filter to the TE before cue detection. Since human reaction time to visual cues approximately ranges from 0.18-0.9s without distractions \cite{fugger_analysis_2000}\cite{hermens_responding_2017}, the critical frequency of the high-pass filter is set to 1Hz in this work. In the current framework, the selection of $\alpha$ and $\beta$ is related to the time constant of DES, which depends on the expected cue period. The time constants for smoothing $\tau_\alpha$ and trend smoothing $\tau_\beta$ are calculated as follows:
\begin{equation}
    \label{time_constant}
    \begin{aligned}
        \tau_\alpha = -\frac{\Delta T}{\ln{(1-\alpha)}} \quad
        \tau_\beta = -\frac{\Delta T}{\ln{(1-\beta)}},
    \end{aligned}
\end{equation}
where $\Delta T$ is the sampling time interval of the data. A basic principle is that $\tau_\alpha$ should be roughly the length of one cue in the setting, and $\tau_\beta$ should be generally shorter than $\tau_\alpha$.

\section{Experiments}
In this section, we first show that TE can be used to quantify information transfer in social interactions. Then we validate the framework in two settings to demonstrate its ability to detect perceptible social cues both temporally and spatially.

\subsection{TE for social interactions - Group-Joining}\label{CongreG8}
To appraise the ability of TE to study social interactions, we first study a group-joining activity. The CongreG8 dataset \cite{yang_dataset_2021} contains 380 full-body motion trials of free-standing conversational groups of three humans and a newcomer who approaches the groups with the intent of joining them. Four participants play a game called \emph{Who's the Spy}. The game contains three group members and one adjudicator (the newcomer). Three group members gather at the centre of the arena, two are given cards with the same item, the third is given a card with a different item. Group members take turns describing the item on their card. Meanwhile, the adjudicator walks around the arena listening to the group members' discussion. Once the adjudicator has identified a spy, they join the group and point out the spy. The interactions are grouped into two categories. When the adjudicator approaches the group and the group members accommodate the adjudicator (e.g., a group member moves to make space), the trial is labelled as \emph{Welcome}. Alternatively, if the group members stand still and ignore the adjudicator, the sample is labelled as \emph{Ignorance}. Due to the high level of freedom within this experiment, participants' behaviours are largely different between each trial resulting in large variance in the sampled data.

We apply the general workflow (\ref{workflow}) to the \emph{Welcome} and \emph{Ignorance} scenarios separately to compute the TE for each trial for two directions (from the adjudicator to the group members, and the reverse) and study the peak values of TE. We set the position of adjudicator or group members as target features depending on the direction of transfer being investigated. We consider the group members as a whole, so a single target/source is defined for all the group members. A re-sampling rate of 10Hz and history length of 10 with a unit time step of 0.1s for the embedding are applied. Since this is a relatively more complex dataset, a Gaussian Emission Variational Autoencoder (VAE) \cite{kingma_auto-encoding_2022} with a two-hidden-layer encoder, a two-hidden-layer decoder and a latent space with 8 nodes is used to model the distributions over targets. 

In this experiment, the adjudicator leads the group-joining event. Depending on the scenario, the responses from the group members are different. For the \emph{Welcome} scenarios, group members should respond more to the perceptible social cues from the adjudicator when compared to the \emph{Ignorance} scenarios. Therefore, we expect to see a difference in peak TE from the adjudicator to the group members because we assume that the peak TE occurs during the joining process when the majority of information transfer from the adjudicator to the group happens. Therefore, we hypothesize that: The peak values of TE from the adjudicator to the group members should have large differences for the \emph{Welcome} and \emph{Ignorance} scenarios, when compared with peak TE in the other direction (group members to adjudicator).

To test this hypothesis, we calculate the peak TE values of each trial for both scenarios and both information transfer directions. Then we conduct a two-sided t-test between the two scenarios for both directions. This test (Table \ref{tab:t-test}) shows a significant difference between \emph{Ignorance} and \emph{Welcome} settings in the adjudicator to group member direction, but not from group members to the adjudicator, supporting our hypothesis. This also supports our proposal that TE can correctly quantify information transfer in social settings.

\begin{table}[!hbt]
\centering
\setlength\belowcaptionskip{-5pt}
\caption{Peak transfer entropy analysis t-test results.}
\label{tab:t-test}
\begin{tabular}{|c|c|}
\hline
\textit{\textbf{Information transfer direction}} & \textit{\textbf{p-value}} \\ \hline
Adjudicator to Group                             & \textbf{0.0008}           \\ \hline
Group to Adjudicator                             & 0.9246                    \\ \hline
\end{tabular}
\vspace{-5mm}
\end{table}

\subsection{Temporal cue detection - Human-Human Handover}

\begin{figure*}[ht!]
\centerline{\includegraphics[width=\linewidth]{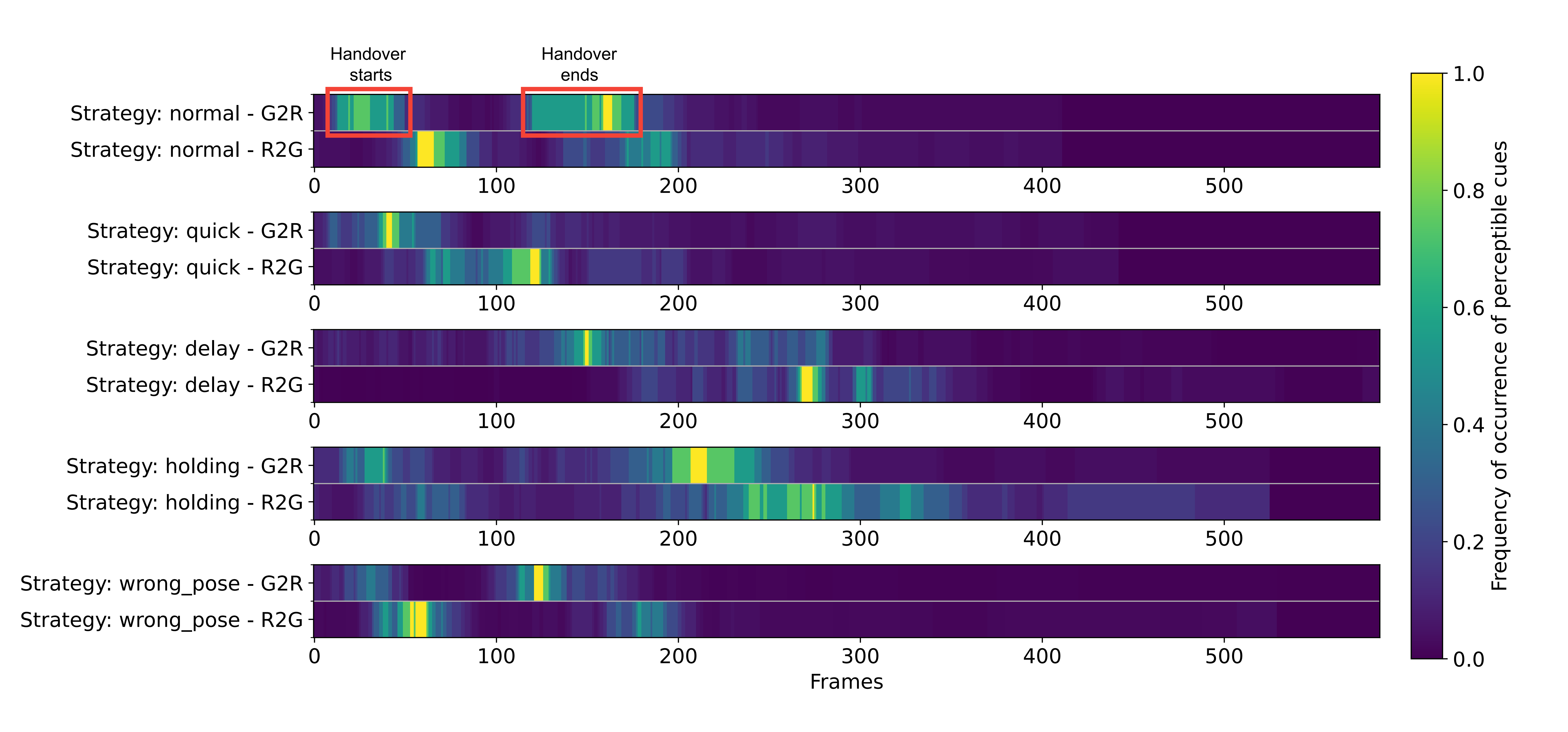}}
\caption{Frequency of occurrence of perceptible cues in the handover task (G - Giver; R - Receiver). The colour indicates the frequency of the occurrence of perceptible cues at each time frame. Cues from the giver always occur before the cues from the receiver. Two high cue count regions are usually identified in each graph indicating the start and end cues of handover respectively (An example is highlighted for the normal scenario). The gap between high cue count regions shows the delay between cues, and the concentration of high cue count regions shows the relative variance in time of the cue. Each graph matches with their corresponding handover strategy.}
\label{HHHO_result}
\Description{Frequency of occurrence of perceptible cues in the handover task plotted along the time frame axis. The colour indicates the frequency of the occurrence of perceptible cues at each time frame. Two high cue count regions are usually identified in each graph indicating the start and end cues of handover respectively. Cues from the giver always occur before the cues from the receiver. The gap between high cue count regions shows the delay between cues, and the concentration of high cue count regions shows the relative variance in time of the cue. Each graph shows a behaviour that matches with their corresponding handover strategy. For example, there is a large gap before the starting cue for the delay strategy, and the gap between starting and ending cue is large for the holding strategy.}
\end{figure*}

We test the framework's ability to identify \emph{when} perceptible social cues occur in a human-human handover scenario. We use a dataset \cite{carfi_multi-sensor_2019} comprising over 1000 recordings collected from 18 right-handed volunteers performing human-human handovers. 6-axis inertial data was acquired from two smartwatches on participants' wrists. During their single-blind experiments, a volunteer (receiver) and an experimenter (giver) form a pair. The two participants start from the diagonal corners of a square experimental area and walk towards the centre of the square to perform handovers. Several strategies are applied during the handovers:
\begin{itemize}
    \item \emph{Normal}: experimenter gives the object in a normal fashion.
    \item \emph{Quick}: experimenter moves their arm faster.
    \item \emph{Delay}: once the volunteer initiates the transferring gesture, the experimenter keeps their arm still ($\approx 2$s) before reaching towards the volunteer to give the object.
    \item \emph{Holding}: experimenter holds the object in place ($\approx 2$s) after both persons have touched it, i.e., they do not release the object once the volunteer has grasped it.
    \item \emph{Wrong pose}: as the volunteer initiates the transfer gesture, the experimenter unexpectedly moves their arm towards the volunteer's left shoulder, an unnatural pose for right-handed volunteers.
\end{itemize}

The original sampling rate of the dataset is approximately $7Hz$, which we interpolated to a frame rate of $115Hz$ to increase the number of window samples to analyse. We used wrist angular velocity data from experiments where the same object, a ball, is used. In this experiment, the magnitude of the triaxial angular velocity is the only feature analysed for perceptible social cues. We use a history length of 4 with a unit time step of 0.14s for the Taken's delay embeddings. These numbers were selected empirically but could be chosen based on model predictive quality. The average human reaction time to detect visual stimuli is approximately 0.18-0.20s \cite{thompson_voluntary_1992}, so 0.4s is a reasonable window to capture human reactions. For cue detection, we set the $\alpha=0.005$ and $\beta=0.01$, which gives us $\tau_\alpha=1.7s$ and $\tau_\beta=0.9s$. We expect that perceptible cues occur when the handover starts and ends with the giver initiating the handover and the receiver reaching in response to their cues \cite{ortenzi_object_2021}.

To demonstrate the ability of the framework, we detect the perceptible cue regions for all the trials using the proposed framework and generate plots that visualise the frequency of occurrence of perceptible cues at each frame for different strategies and information transfer directions. The results are shown in Fig. \ref{HHHO_result}. We trim the data so the frame consistently starts when the distance between the two participants is 1m. In a regular handover scenario, two main social cues should occur at the start and end of a handover respectively \cite{ortenzi_object_2021}. We also expect the cues from a giver to be followed by the cues from the receiver. This is visible in the \emph{normal} strategy result, with two high cue count regions in both directions. The first and second regions correspond to the start and end of the handover respectively. We can see that high cue count regions in the receiver-to-giver (R2G) direction follow the giver-to-receiver (G2R) direction. Using the \emph{normal} strategy as a baseline for comparison, we can also analyse the other scenarios. In the \emph{quick} scenario, due to the faster movements, the cues occur and end faster when compared to the \emph{normal} strategy; i.e, the spacing between cue counts is comparatively shorter. The fast movement of the giver could also confuse the receiver, resulting in a wider spread of the receiver's reaction cue. In the \emph{delay} scenario, the participant was not given instructions on how long to delay during the experiment, thus the time of the starting cue of the handover is widely distributed. We observe a large delay before the first high cue count region appears, and the cues are more widely distributed. For the \emph{holding} strategy, the time gap between the starting and ending cue should be longer compared to \emph{normal}. This larger gap is observed between the first and second high cue count regions in both directions. For the \emph{wrong pose} scenario, the wrong pose of the giver could confuse the receiver, resulting in weaker responses from the receiver. Therefore, the plots appear to be similar to the \emph{normal} scenario. However, we observe the first high cue count region in the G2R direction has a lower cue occurrence frequency compared to \emph{normal}. This analysis demonstrates that the proposed framework can identify \emph{when} perceptible social cues occur. Additionally, interesting behaviour patterns can be extracted from the TE analysis, which can be useful for downstream anthropomorphic and personalized robot interaction design.

\subsection{Spatial cue detection - Person-Following}

We next apply the proposed framework to \emph{spatially} analyse perceptible social cue transfer in a socially-aware navigation scenario requiring substantial non-verbal communication.

\begin{figure}[!ht]
\centering
  \begin{minipage}[b]{.22\textwidth}
  \subfloat
    [][]{\label{bird_view}\includegraphics[width=\textwidth,height=5.9cm]{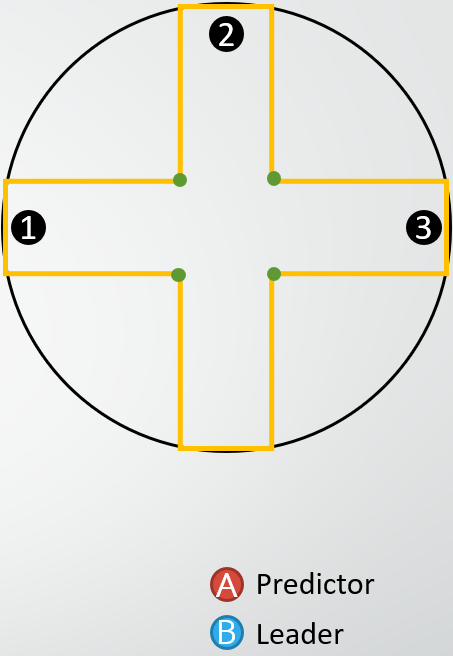}}
  \end{minipage}
\begin{minipage}[b]{.22\textwidth}
\centering
\subfloat
  [][] {\label{side_view}\includegraphics[height=2.5cm]{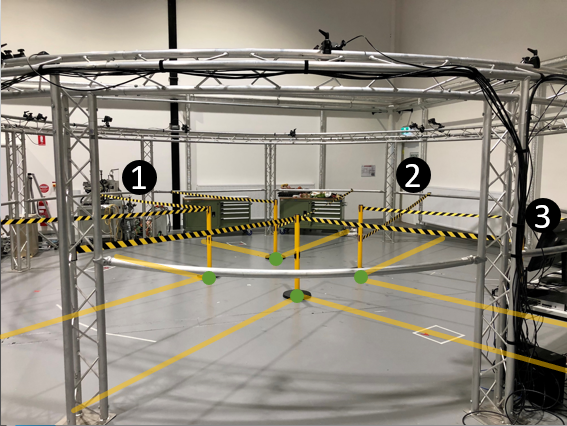}}
\vspace{-0.4cm}
\subfloat
  [][]{\label{sketch}\includegraphics[height=3.0cm]{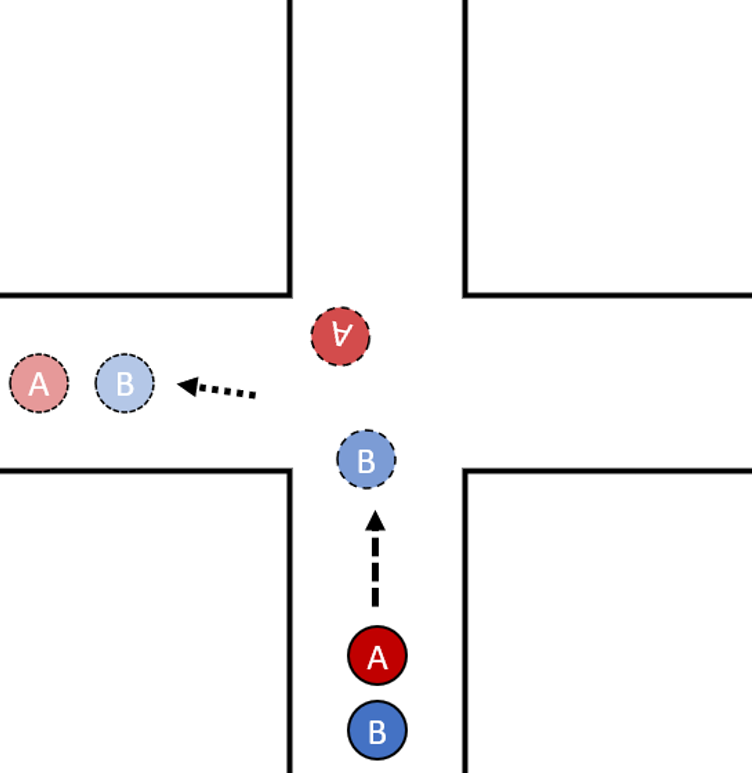}}
  \vspace{-5mm}
\end{minipage}
\caption{Person-following task design. (a) An illustration of the starting position and formation. (b) A side view of the experimental arena. 
(c) An example leader-predictor front-following task (turning left scenario).}
\label{fig:PF_design}
\Description{The design for the front-following experiment. The Vicon motion capture camera arena is a circle, and the simulated indoor cross intersection is built inside the circle. There is one entrance and three ending points labelled as 1, 2 and 3 locating at the end of three branches of the intersection. The participants start slightly outside the circle in a front and back formation (Predictor stays in front of the leader) and facing the entrance of the intersection.}
\end{figure}

To maximise the exchange of perceptible social cues, we designed a leader-predictor front-following task simulating a person-following robot scenario (Fig. \ref{sketch}) in a simulated intersection assembled using retractable barriers in a Vicon motion capture arena (Fig. \ref{side_view}). The arena has a diameter of approximately 5.0m, and the junction size is a 1.6m$\times$1.6m square to permit comfortable side-by-side walking \cite{noauthor_two_nodate}. The scenario involves two human participants: the predictor A and the leader B. At the beginning of each trial, B is asked to secretly select one of the destinations shown in Fig. \ref{bird_view}. A is asked to actively stay in front of B and attempt to reach the unknown destination in advance of B. The starting position and formation are illustrated in Fig. \ref{bird_view}. This design avoids potential bias where the initial pose of participants might influence their preference after entering the intersection. When each trial begins, B walks naturally towards the selected destination, treating A as an unknown pedestrian (i.e., keeping a comfortable distance, no direct communication, but still following general social norms such as adjusting speed to avoid collision). After entering the intersection (the black circle in Fig. \ref{bird_view}), A can move freely, provided they do not interfere with B's progress. Direct communication channels such as speaking or hand signals are forbidden, forcing participants to communicate via more subtle motion cues. This `front-following' task involves high levels of observation, prediction and collaboration, providing an interesting setting to analyse the use of social cues. Seven pairs (14 individuals aged 19 to 31, 3 females, 11 males) were recruited from the Monash University Clayton campus. Participants provided consent at the start of the experiment \footnote{Approved by the Monash University ethics committee. Project ID: 33090}. 

\begin{figure*}[ht!]
\centerline{\includegraphics[width=0.8\linewidth]{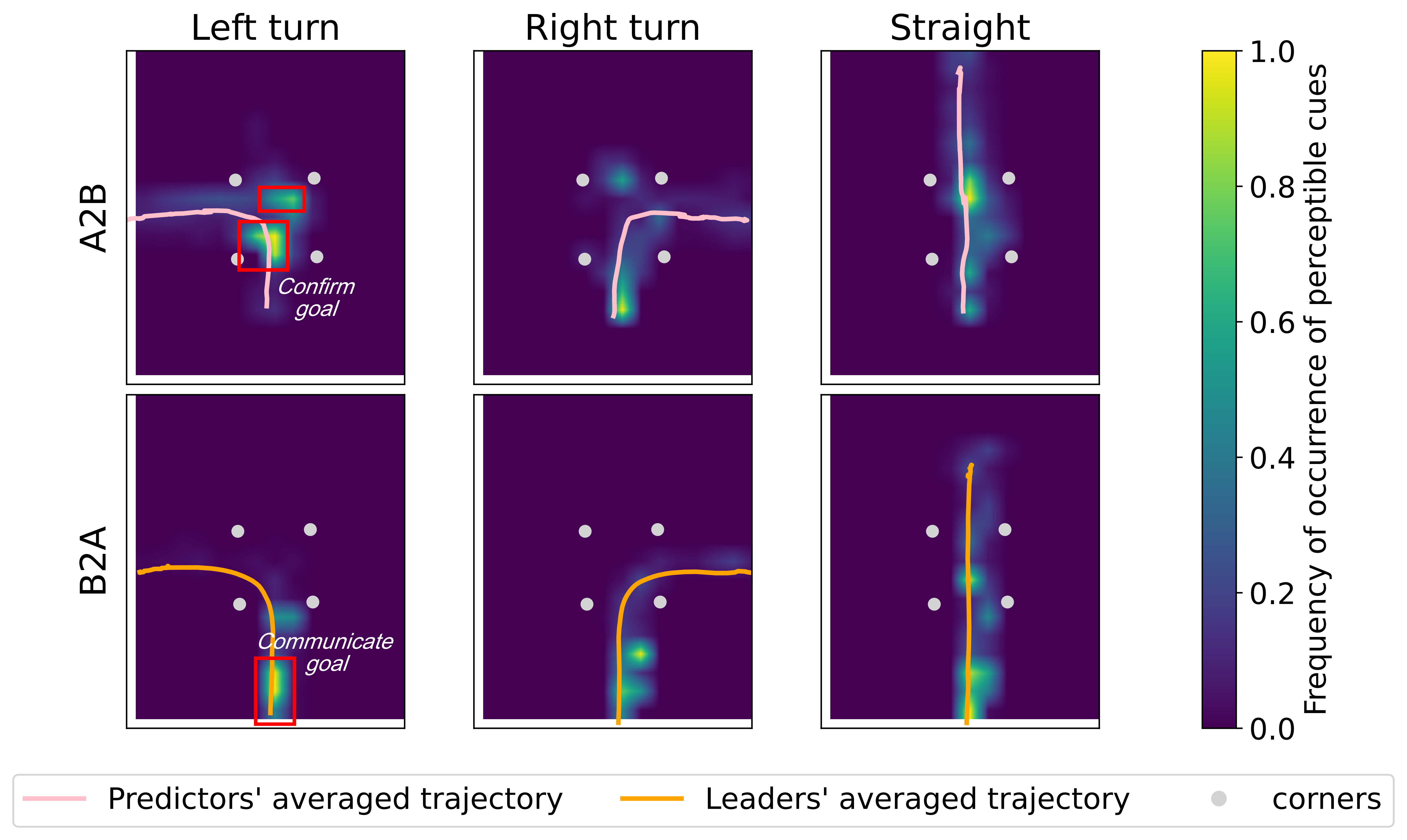}}
\setlength\abovecaptionskip{-0.5pt}
\caption{Frequency of occurrence (cue count) of perceptible cues in the front-following task (A - Predictor; B - Leader). The colour indicates the frequency of the occurrence of perceptible cues at a physical location. Cues sent from the leader to the predictor to communicate the goal usually occur further away from the intersection, while cues sent from the predictor to the leader to confirm direction usually occur closer to or inside the intersection. High cue count regions are annotated in the left turn scenario as an example. This pattern in physical space is consistent for the three scenarios.}
\label{IFF_heatmap}
\Description{Frequency of occurrence (cue count) of perceptible cues in the front-following task plotted in 2-dimensional heatmaps. The colour indicates the frequency of the occurrence of perceptible cues at a physical location. It shows that the cues sent from the leader to the predictor to communicate the goal usually occur further away from the intersection, while cues sent from the predictor to the leader to confirm direction usually occur closer to or inside the intersection. This pattern in physical space is consistent for the three scenarios.}
\end{figure*}

The Vicon system provides a default frame rate of around $200$ Hz. Vicon markers on a cap and a belt are used to track A and B's head and hip pose, with the centre position and the head and hip orientation recorded for each participant.\footnote{The data is publicly available at \href{https://doi.org/10.26180/24719034.v1}{https://doi.org/10.26180/24719034.v1}} Head and hip velocity are post-calculated using the positional information. Orientation and velocity vectors are projected onto the \textit{xy} plane and then normalised. To apply the proposed framework, we take position, velocity, head and hip orientation as potential sources of cues and set each participant's position as the target. A variational autoencoder model is fit to the target using embeddings with a history length of 4 and a unit time step of 0.1s. For cue detection, we set the $\alpha=0.01$ and $\beta=0.05$, which gives us time constants $\tau_\alpha=0.5s$ and $\tau_\beta=0.1s$.  Compared to the handover task, the cues in this experiment can be quicker, subtler and noisier. Our hypothesis is that perceptible cues appear in locations where a decision needs to be made or communicated about travel direction.

Similar to the handover experiment, we detect perceptible cue regions for all the trials using the proposed framework and generate plots that visualise the perceptible cue frequency at each spatial location in a 2-dimensional grid space representing the experiment arena. The results are shown in Fig. \ref{IFF_heatmap}. We observe that for the leader to predictor (B2A) direction, most high-frequency cue regions appear in the region leading to the intersection. For the predictor to leader (A2B) direction, high-frequency cue regions often appear inside or slightly before the intersection. This outcome is consistent with the leader's role in the front-following task, who first transfers cues to the predictor (B2A) prior to entering the intersection.  This communication is usually followed by cues confirming the travel direction sent by the predictor (A2B) before leaving the intersection. This pattern of cue communication in physical space is clearly shown in Fig. \ref{IFF_heatmap}. Although in theory, we would expect the perceptible cue regions to be similar for tuning left and right scenarios, the results show that they have noticeable differences. 
Most predictors (A) preferred to stay in the top left corner of the intersection while waiting and observing the leader (B) during the experiment \footnote{This experiment is done in Australia where the norm is to keep left on sidewalks.}. This matches with post-experiment survey results that showed 42.9\% of the predictors usually prefer to walk on the left side of the road or a corridor (28.6\% right and 28.6\% no preference). Therefore, physical space occupancy could be an influential factor for information transfer. This highlights the potential of the framework to capture information about cultural navigation biases.

The analysis above shows the proposed framework can identify \emph{where} perceptible social cues occur.

\section{Discussion}\label{Disucssion}

Our experiments have demonstrated the capability of the proposed framework. In addition to detecting perceptible social cues, the proposed framework can be potentially used to conduct further analysis. For instance, it has good scalability in the sense that it allows us to analyse detailed cues related to individual motion features, or combined cues if we consider all motions together. This scalability can be used to zero in on the source of cues embedded in motion. In addition, as shown in \ref{CongreG8}, the magnitude of TE potentially represents the strength of information transfer. This means it is possible to compare the strength of each cue in a setting. 

While our experiments have focused on human interactions, we believe the proposed framework is broadly applicable in the field of HRI. Studying human-human cues can help robot designers identify suitable sources of information or cues for more natural human-robot interaction. The proposed framework could also be used to analyse social cues or information transfer during collaboration between humans and robots, for example in handover between humans and manipulators. In the context of socially-aware navigation, we could design navigation algorithms to control the strength of information transfer. For example, we could intentionally reduce the influence on pedestrians during socially-aware navigation by minimising TE, thus minimising the influence of a robot on pedestrians. A robot could also take actions to increase information transfer to convey more information and potentially influence pedestrian motion if needed. This approach is potentially more flexible, adaptive and inclusive than current methods \cite{tranberg_hansen_adaptive_2009}\cite{mi_system_2016}\cite{hu_design_2014}, which are designed based on Hall's proxemics theory and the concept of social zones \cite{hall_hidden_1966} that has been criticised for not considering the diverse range of social norms demonstrated by humans \cite{rios-martinez_proxemics_2015}.

\section{Conclusions and Future Work}
In this paper, we propose a framework for analysing perceptible social cue information transfer using transfer entropy. We have used a group-joining experiment to show the ability of TE to analyse information exchange during social interactions. We have applied the proposed framework to two unique settings, namely: object-handover and person-front-following, and demonstrated its capability of identifying the temporal and spatial occurrence of perceptible social cues. The proposed framework can be used for analysing cue information transfer of human-human interactions to identify cues of potential interest to robot designers, but could also be applied to analyse social cue transfer during human-robot collaborations.

Extending the proposed framework to real-time human-robot collaboration is a particularly exciting area of future work. This would require technical advancements and modelling choices to allow for distributions to be learned online in new settings, but this could be simplified if pre-trained distributions are obtained from previously observed settings. Further user studies are needed to determine if similar social cues can be elicited or generated when a robot is a participant in an interaction, and these studies are an exciting next step for us. We plan to assess the capability of the proposed framework in less controlled settings and compare it with other existing cue detection methods.


\begin{acks}
We are grateful to Dr. Wesley Chan for his assistance at the early stage of this research and Mr. Brandon Johns for technical support with the Vicon system.
\end{acks}

\bibliographystyle{ACM-Reference-Format}
\bibliography{references}


\begin{thebibliography}{46}


\ifx \showCODEN    \undefined \def \showCODEN     #1{\unskip}     \fi
\ifx \showDOI      \undefined \def \showDOI       #1{#1}\fi
\ifx \showISBNx    \undefined \def \showISBNx     #1{\unskip}     \fi
\ifx \showISBNxiii \undefined \def \showISBNxiii  #1{\unskip}     \fi
\ifx \showISSN     \undefined \def \showISSN      #1{\unskip}     \fi
\ifx \showLCCN     \undefined \def \showLCCN      #1{\unskip}     \fi
\ifx \shownote     \undefined \def \shownote      #1{#1}          \fi
\ifx \showarticletitle \undefined \def \showarticletitle #1{#1}   \fi
\ifx \showURL      \undefined \def \showURL       {\relax}        \fi
\providecommand\bibfield[2]{#2}
\providecommand\bibinfo[2]{#2}
\providecommand\natexlab[1]{#1}
\providecommand\showeprint[2][]{arXiv:#2}

\bibitem[Baek et~al\mbox{.}(2005)]%
        {baek_transfer_2005}
\bibfield{author}{\bibinfo{person}{Seung~Ki Baek}, \bibinfo{person}{Woo-Sung Jung}, \bibinfo{person}{Okyu Kwon}, {and} \bibinfo{person}{Hie-Tae Moon}.} \bibinfo{year}{2005}\natexlab{}.
\newblock \showarticletitle{Transfer Entropy Analysis of the Stock Market}. In \bibinfo{booktitle}{\emph{arXiv}}. \bibinfo{publisher}{{arXiv}}, \bibinfo{address}{arXiv}.
\newblock
\urldef\tempurl%
\url{https://doi.org/10.48550/arXiv.physics/0509014}
\showDOI{\tempurl}
\showeprint[arxiv]{physics/0509014}


\bibitem[Baur et~al\mbox{.}(2013)]%
        {hutchison_nova_2013}
\bibfield{author}{\bibinfo{person}{Tobias Baur}, \bibinfo{person}{Ionut Damian}, \bibinfo{person}{Florian Lingenfelser}, \bibinfo{person}{Johannes Wagner}, {and} \bibinfo{person}{Elisabeth Andr{\'e}}.} \bibinfo{year}{2013}\natexlab{}.
\newblock \showarticletitle{NovA: Automated Analysis of Nonverbal Signals in Social Interactions}. In \bibinfo{booktitle}{\emph{Human Behavior Understanding}}, \bibfield{editor}{\bibinfo{person}{Albert~Ali Salah}, \bibinfo{person}{Hayley Hung}, \bibinfo{person}{Oya Aran}, {and} \bibinfo{person}{Hatice Gunes}} (Eds.). \bibinfo{publisher}{Springer International Publishing}, \bibinfo{address}{Cham}, \bibinfo{pages}{160--171}.
\newblock
\showISBNx{978-3-319-02714-2}


\bibitem[Berger et~al\mbox{.}(2014)]%
        {berger_transfer_2014}
\bibfield{author}{\bibinfo{person}{Erik Berger}, \bibinfo{person}{David Müller}, \bibinfo{person}{David Vogt}, \bibinfo{person}{Bernhard Jung}, {and} \bibinfo{person}{Heni Ben~Amor}.} \bibinfo{year}{2014}\natexlab{}.
\newblock \showarticletitle{Transfer entropy for feature extraction in physical human-robot interaction: Detecting perturbations from low-cost sensors}. In \bibinfo{booktitle}{\emph{2014 IEEE-RAS International Conference on Humanoid Robots}} (Madrid, Spain). \bibinfo{publisher}{IEEE}, \bibinfo{address}{New York, NY, USA}, \bibinfo{pages}{829--834}.
\newblock
\urldef\tempurl%
\url{https://doi.org/10.1109/HUMANOIDS.2014.7041459}
\showDOI{\tempurl}


\bibitem[Bossomaier et~al\mbox{.}(2016)]%
        {bossomaier_transfer_2016}
\bibfield{author}{\bibinfo{person}{Terry Bossomaier}, \bibinfo{person}{Lionel Barnett}, \bibinfo{person}{Michael Harr{\'e}}, {and} \bibinfo{person}{Joseph~T. Lizier}.} \bibinfo{year}{2016}\natexlab{}.
\newblock \showarticletitle{Transfer Entropy}.
\newblock In \bibinfo{booktitle}{\emph{An Introduction to Transfer Entropy: Information Flow in Complex Systems}}. \bibinfo{publisher}{Springer International Publishing}, \bibinfo{address}{Cham}, \bibinfo{pages}{65--95}.
\newblock
\showISBNx{978-3-319-43222-9}
\urldef\tempurl%
\url{https://doi.org/10.1007/978-3-319-43222-9_4}
\showDOI{\tempurl}


\bibitem[Bousmalis et~al\mbox{.}(2 01)]%
        {bousmalis_towards_2013}
\bibfield{author}{\bibinfo{person}{Konstantinos Bousmalis}, \bibinfo{person}{Marc Mehu}, {and} \bibinfo{person}{Maja Pantic}.} \bibinfo{year}{2013-02-01}\natexlab{}.
\newblock \showarticletitle{Towards the automatic detection of spontaneous agreement and disagreement based on nonverbal behaviour: A survey of related cues, databases, and tools}.
\newblock \bibinfo{journal}{\emph{Image and Vision Computing}} \bibinfo{volume}{31}, \bibinfo{number}{2} (\bibinfo{year}{2013-02-01}), \bibinfo{pages}{203--221}.
\newblock
\showISSN{0262-8856}
\urldef\tempurl%
\url{https://doi.org/10.1016/j.imavis.2012.07.003}
\showDOI{\tempurl}


\bibitem[Bremers et~al\mbox{.}(2023)]%
        {bremers_using_2023}
\bibfield{author}{\bibinfo{person}{Alexandra Bremers}, \bibinfo{person}{Alexandria Pabst}, \bibinfo{person}{Maria~Teresa Parreira}, {and} \bibinfo{person}{Wendy Ju}.} \bibinfo{year}{2023}\natexlab{}.
\newblock \showarticletitle{Using Social Cues to Recognize Task Failures for {HRI}: A Review of Current Research and Future Directions}. In \bibinfo{booktitle}{\emph{arXiv}}. \bibinfo{publisher}{{arXiv}}, \bibinfo{address}{arXiv}.
\newblock
\urldef\tempurl%
\url{https://doi.org/10.48550/arXiv.2301.11972}
\showDOI{\tempurl}
\showeprint[arxiv]{2301.11972 [cs]}


\bibitem[Busch et~al\mbox{.}(2017)]%
        {busch_learning_2017}
\bibfield{author}{\bibinfo{person}{Baptiste Busch}, \bibinfo{person}{Jonathan Grizou}, \bibinfo{person}{Manuel Lopes}, {and} \bibinfo{person}{Freek Stulp}.} \bibinfo{year}{2017}\natexlab{}.
\newblock \showarticletitle{Learning Legible Motion from Human–Robot Interactions}.
\newblock \bibinfo{journal}{\emph{Int J of Soc Robotics}} \bibinfo{volume}{9}, \bibinfo{number}{5} (\bibinfo{year}{2017}), \bibinfo{pages}{765--779}.
\newblock
\showISSN{1875-4805}
\urldef\tempurl%
\url{https://doi.org/10.1007/s12369-017-0400-4}
\showDOI{\tempurl}


\bibitem[Carfì et~al\mbox{.}(2019)]%
        {carfi_multi-sensor_2019}
\bibfield{author}{\bibinfo{person}{Alessandro Carfì}, \bibinfo{person}{Francesco Foglino}, \bibinfo{person}{Barbara Bruno}, {and} \bibinfo{person}{Fulvio Mastrogiovanni}.} \bibinfo{year}{2019}\natexlab{}.
\newblock \showarticletitle{A multi-sensor dataset of human-human handover}.
\newblock \bibinfo{journal}{\emph{Data in Brief}}  \bibinfo{volume}{22} (\bibinfo{year}{2019}), \bibinfo{pages}{109--117}.
\newblock
\showISSN{2352-3409}
\urldef\tempurl%
\url{https://doi.org/10.1016/j.dib.2018.11.110}
\showDOI{\tempurl}


\bibitem[Doss(2023)]%
        {noauthor_two_nodate}
\bibfield{author}{\bibinfo{person}{Minnie Doss}.} \bibinfo{year}{2023}\natexlab{}.
\newblock \bibinfo{title}{Two Lane Corridor Dimensions \& Drawings {\textbar} Dimensions.com}.
\newblock
\newblock


\bibitem[Dragan et~al\mbox{.}(2013)]%
        {dragan_legibility_2013}
\bibfield{author}{\bibinfo{person}{Anca~D. Dragan}, \bibinfo{person}{Kenton~C.T. Lee}, {and} \bibinfo{person}{Siddhartha~S. Srinivasa}.} \bibinfo{year}{2013}\natexlab{}.
\newblock \showarticletitle{Legibility and predictability of robot motion}. In \bibinfo{booktitle}{\emph{2013 8th ACM/IEEE International Conference on Human-Robot Interaction (HRI)}} (Tokyo, Japan). \bibinfo{publisher}{IEEE}, \bibinfo{address}{New York, NY, USA}, \bibinfo{pages}{301--308}.
\newblock
\urldef\tempurl%
\url{https://doi.org/10.1109/HRI.2013.6483603}
\showDOI{\tempurl}


\bibitem[Escobedo et~al\mbox{.}(2014)]%
        {escobedo_using_2014}
\bibfield{author}{\bibinfo{person}{Arturo Escobedo}, \bibinfo{person}{Anne Spalanzani}, {and} \bibinfo{person}{Christian Laugier}.} \bibinfo{year}{2014}\natexlab{}.
\newblock \showarticletitle{Using social cues to estimate possible destinations when driving a robotic wheelchair}. In \bibinfo{booktitle}{\emph{2014 IEEE/RSJ International Conference on Intelligent Robots and Systems}} (Chicago, IL, USA). \bibinfo{publisher}{IEEE}, \bibinfo{address}{New York, NY, USA}, \bibinfo{pages}{3299--3304}.
\newblock
\urldef\tempurl%
\url{https://doi.org/10.1109/IROS.2014.6943021}
\showDOI{\tempurl}


\bibitem[Fugger et~al\mbox{.}(1 01)]%
        {fugger_analysis_2000}
\bibfield{author}{\bibinfo{person}{Thomas Fugger}, \bibinfo{person}{Bryan Randles}, \bibinfo{person}{Anthony Stein}, \bibinfo{person}{William Whiting}, {and} \bibinfo{person}{Brian Gallagher}.} \bibinfo{year}{2000-01-01}\natexlab{}.
\newblock \showarticletitle{Analysis of Pedestrian Gait and Perception-Reaction at Signal-Controlled Crosswalk Intersections}.
\newblock \bibinfo{journal}{\emph{Transportation Research Record}}  \bibinfo{volume}{1705} (\bibinfo{year}{2000-01-01}), \bibinfo{pages}{20--25}.
\newblock
\urldef\tempurl%
\url{https://doi.org/10.3141/1705-04}
\showDOI{\tempurl}


\bibitem[Gardner~Jr.(1985)]%
        {gardner_jr_exponential_1985}
\bibfield{author}{\bibinfo{person}{Everette~S. Gardner~Jr.}} \bibinfo{year}{1985}\natexlab{}.
\newblock \showarticletitle{Exponential smoothing: The state of the art}.
\newblock \bibinfo{journal}{\emph{Journal of Forecasting}} \bibinfo{volume}{4}, \bibinfo{number}{1} (\bibinfo{year}{1985}), \bibinfo{pages}{1--28}.
\newblock
\showISSN{1099-131X}
\urldef\tempurl%
\url{https://doi.org/10.1002/for.3980040103}
\showDOI{\tempurl}


\bibitem[Granger(1969)]%
        {granger_investigating_1969}
\bibfield{author}{\bibinfo{person}{C.~W.~J. Granger}.} \bibinfo{year}{1969}\natexlab{}.
\newblock \showarticletitle{Investigating Causal Relations by Econometric Models and Cross-spectral Methods}.
\newblock \bibinfo{journal}{\emph{Econometrica}} \bibinfo{volume}{37}, \bibinfo{number}{3} (\bibinfo{year}{1969}), \bibinfo{pages}{424--438}.
\newblock
\showISSN{0012-9682}
\urldef\tempurl%
\url{https://doi.org/10.2307/1912791}
\showDOI{\tempurl}


\bibitem[Green and Swets(1966)]%
        {green_signal_1966}
\bibfield{author}{\bibinfo{person}{David~M. Green} {and} \bibinfo{person}{John~A. Swets}.} \bibinfo{year}{1966}\natexlab{}.
\newblock \bibinfo{booktitle}{\emph{Signal detection theory and psychophysics}}.
\newblock \bibinfo{publisher}{John Wiley}, \bibinfo{address}{Oxford, England}.
\newblock


\bibitem[Hall(1966)]%
        {hall_hidden_1966}
\bibfield{author}{\bibinfo{person}{Edward~T. Hall}.} \bibinfo{year}{1966}\natexlab{}.
\newblock \bibinfo{booktitle}{\emph{The hidden dimension: man’s use of space in public and private}}.
\newblock \bibinfo{publisher}{The Bodley Head Ltd}, \bibinfo{address}{London}.
\newblock


\bibitem[He and Shang(2017)]%
        {he_comparison_2017}
\bibfield{author}{\bibinfo{person}{Jiayi He} {and} \bibinfo{person}{Pengjian Shang}.} \bibinfo{year}{2017}\natexlab{}.
\newblock \showarticletitle{Comparison of transfer entropy methods for financial time series}.
\newblock \bibinfo{journal}{\emph{Physica A: Statistical Mechanics and its Applications}}  \bibinfo{volume}{482} (\bibinfo{year}{2017}), \bibinfo{pages}{772--785}.
\newblock
\showISSN{0378-4371}
\urldef\tempurl%
\url{https://doi.org/10.1016/j.physa.2017.04.089}
\showDOI{\tempurl}


\bibitem[Hermens et~al\mbox{.}(1 01)]%
        {hermens_responding_2017}
\bibfield{author}{\bibinfo{person}{Frouke Hermens}, \bibinfo{person}{Markus Bindemann}, {and} \bibinfo{person}{A. Mike~Burton}.} \bibinfo{year}{2017-01-01}\natexlab{}.
\newblock \showarticletitle{Responding to social and symbolic extrafoveal cues: cue shape trumps biological relevance}.
\newblock \bibinfo{journal}{\emph{Psychological Research}} \bibinfo{volume}{81}, \bibinfo{number}{1} (\bibinfo{year}{2017-01-01}), \bibinfo{pages}{24--42}.
\newblock
\showISSN{1430-2772}
\urldef\tempurl%
\url{https://doi.org/10.1007/s00426-015-0733-2}
\showDOI{\tempurl}


\bibitem[Hetherington et~al\mbox{.}(2021)]%
        {hetherington_hey_2021}
\bibfield{author}{\bibinfo{person}{Nicholas~J. Hetherington}, \bibinfo{person}{Elizabeth~A. Croft}, {and} \bibinfo{person}{H.~F.~Machiel Van~der Loos}.} \bibinfo{year}{2021}\natexlab{}.
\newblock \showarticletitle{Hey Robot, Which Way Are You Going? Nonverbal Motion Legibility Cues for Human-Robot Spatial Interaction}.
\newblock \bibinfo{journal}{\emph{{IEEE} Robot. Autom. Lett.}} \bibinfo{volume}{6}, \bibinfo{number}{3} (\bibinfo{year}{2021}), \bibinfo{pages}{5010--5015}.
\newblock
\showISSN{2377-3766, 2377-3774}
\urldef\tempurl%
\url{https://doi.org/10.1109/LRA.2021.3068708}
\showDOI{\tempurl}
\showeprint[arxiv]{2104.02275}


\bibitem[Hu et~al\mbox{.}(4 04)]%
        {hu_design_2014}
\bibfield{author}{\bibinfo{person}{Jwu-Sheng Hu}, \bibinfo{person}{Jyun-Ji Wang}, {and} \bibinfo{person}{Daniel~Minare Ho}.} \bibinfo{year}{2014-04}\natexlab{}.
\newblock \showarticletitle{Design of Sensing System and Anticipative Behavior for Human Following of Mobile Robots}.
\newblock \bibinfo{journal}{\emph{{IEEE} Trans. on Industrial Electronics}} \bibinfo{volume}{61}, \bibinfo{number}{4} (\bibinfo{year}{2014-04}), \bibinfo{pages}{1916--1927}.
\newblock
\showISSN{1557-9948}
\urldef\tempurl%
\url{https://doi.org/10.1109/TIE.2013.2262758}
\showDOI{\tempurl}


\bibitem[Jaques et~al\mbox{.}(2019)]%
        {jaques_social_2019}
\bibfield{author}{\bibinfo{person}{Natasha Jaques}, \bibinfo{person}{Angeliki Lazaridou}, \bibinfo{person}{Edward Hughes}, \bibinfo{person}{Caglar Gulcehre}, \bibinfo{person}{Pedro Ortega}, \bibinfo{person}{Dj Strouse}, \bibinfo{person}{Joel~Z. Leibo}, {and} \bibinfo{person}{Nando De~Freitas}.} \bibinfo{year}{2019}\natexlab{}.
\newblock \showarticletitle{Social Influence as Intrinsic Motivation for Multi-Agent Deep Reinforcement Learning}. In \bibinfo{booktitle}{\emph{Proceedings of the 36th International Conference on Machine Learning}} \emph{(\bibinfo{series}{Proceedings of Machine Learning Research}, Vol.~\bibinfo{volume}{97})}, \bibfield{editor}{\bibinfo{person}{Kamalika Chaudhuri} {and} \bibinfo{person}{Ruslan Salakhutdinov}} (Eds.). \bibinfo{publisher}{PMLR}, \bibinfo{address}{CA, USA}, \bibinfo{pages}{3040--3049}.
\newblock


\bibitem[Kingma and Welling(2022)]%
        {kingma_auto-encoding_2022}
\bibfield{author}{\bibinfo{person}{Diederik~P. Kingma} {and} \bibinfo{person}{Max Welling}.} \bibinfo{year}{2022}\natexlab{}.
\newblock \showarticletitle{Auto-Encoding Variational Bayes}. In \bibinfo{booktitle}{\emph{arXiv}}. \bibinfo{publisher}{arXiv}, \bibinfo{address}{arXiv}.
\newblock
\urldef\tempurl%
\url{https://doi.org/10.48550/arXiv.1312.6114}
\showDOI{\tempurl}
\showeprint[arxiv]{1312.6114 [cs, stat]}


\bibitem[Kistler et~al\mbox{.}(7 01)]%
        {kistler_natural_2012}
\bibfield{author}{\bibinfo{person}{Felix Kistler}, \bibinfo{person}{Birgit Endrass}, \bibinfo{person}{Ionut Damian}, \bibinfo{person}{Chi~Tai Dang}, {and} \bibinfo{person}{Elisabeth André}.} \bibinfo{year}{2012-07-01}\natexlab{}.
\newblock \showarticletitle{Natural interaction with culturally adaptive virtual characters}.
\newblock \bibinfo{journal}{\emph{Journal on Multimodal User Interfaces}} \bibinfo{volume}{6}, \bibinfo{number}{1} (\bibinfo{year}{2012-07-01}), \bibinfo{pages}{39--47}.
\newblock
\showISSN{1783-8738}
\urldef\tempurl%
\url{https://doi.org/10.1007/s12193-011-0087-z}
\showDOI{\tempurl}


\bibitem[Klyubin et~al\mbox{.}(2005)]%
        {klyubin_empowerment_2005}
\bibfield{author}{\bibinfo{person}{A.S. Klyubin}, \bibinfo{person}{D. Polani}, {and} \bibinfo{person}{C.L. Nehaniv}.} \bibinfo{year}{2005}\natexlab{}.
\newblock \showarticletitle{Empowerment: a universal agent-centric measure of control}. In \bibinfo{booktitle}{\emph{2005 IEEE Congress on Evolutionary Computation}} (Edinburgh, UK), Vol.~\bibinfo{volume}{1}. \bibinfo{publisher}{IEEE}, \bibinfo{address}{New York, NY, USA}, \bibinfo{pages}{128--135 Vol.1}.
\newblock
\urldef\tempurl%
\url{https://doi.org/10.1109/CEC.2005.1554676}
\showDOI{\tempurl}


\bibitem[Leibo et~al\mbox{.}(2017)]%
        {leibo_multi-agent_2017}
\bibfield{author}{\bibinfo{person}{Joel~Z. Leibo}, \bibinfo{person}{Vinicius Zambaldi}, \bibinfo{person}{Marc Lanctot}, \bibinfo{person}{Janusz Marecki}, {and} \bibinfo{person}{Thore Graepel}.} \bibinfo{year}{2017}\natexlab{}.
\newblock \showarticletitle{Multi-Agent Reinforcement Learning in Sequential Social Dilemmas}. In \bibinfo{booktitle}{\emph{Proceedings of the 16th Conference on Autonomous Agents and MultiAgent Systems}} (S\~{a}o Paulo, Brazil) \emph{(\bibinfo{series}{AAMAS '17})}. \bibinfo{publisher}{International Foundation for Autonomous Agents and Multiagent Systems}, \bibinfo{address}{Richland, SC}, \bibinfo{pages}{464–473}.
\newblock


\bibitem[Lichtenthäler et~al\mbox{.}(2012)]%
        {lichtenthaler_influence_2012}
\bibfield{author}{\bibinfo{person}{Christina Lichtenthäler}, \bibinfo{person}{Tamara Lorenzy}, {and} \bibinfo{person}{Alexandra Kirsch}.} \bibinfo{year}{2012}\natexlab{}.
\newblock \showarticletitle{Influence of legibility on perceived safety in a virtual human-robot path crossing task}. In \bibinfo{booktitle}{\emph{2012 IEEE RO-MAN: The 21st IEEE International Symposium on Robot and Human Interactive Communication}} (Paris, France). \bibinfo{publisher}{IEEE}, \bibinfo{address}{New York, NY, USA}, \bibinfo{pages}{676--681}.
\newblock
\urldef\tempurl%
\url{https://doi.org/10.1109/ROMAN.2012.6343829}
\showDOI{\tempurl}


\bibitem[Lizier(2014)]%
        {lizier_jidt_2014}
\bibfield{author}{\bibinfo{person}{Joseph~T. Lizier}.} \bibinfo{year}{2014}\natexlab{}.
\newblock \showarticletitle{JIDT: An Information-Theoretic Toolkit for Studying the Dynamics of Complex Systems}.
\newblock \bibinfo{journal}{\emph{Frontiers in Robotics and {AI}}}  \bibinfo{volume}{1} (\bibinfo{year}{2014}).
\newblock
\showISSN{2296-9144}
\urldef\tempurl%
\url{https://doi.org/10.3389/frobt.2014.00011}
\showDOI{\tempurl}
\showeprint[arxiv]{1408.3270 [nlin, physics:physics]}


\bibitem[Mi et~al\mbox{.}(2016)]%
        {mi_system_2016}
\bibfield{author}{\bibinfo{person}{Weiming Mi}, \bibinfo{person}{Xiaozhe Wang}, \bibinfo{person}{Ping Ren}, {and} \bibinfo{person}{Chenyue Hou}.} \bibinfo{year}{2016}\natexlab{}.
\newblock \showarticletitle{A System for an Anticipative Front Human Following Robot}. In \bibinfo{booktitle}{\emph{Proceedings of the International Conference on Artificial Intelligence and Robotics and the International Conference on Automation, Control and Robotics Engineering}} (Kitakyushu, Japan) \emph{(\bibinfo{series}{ICAIR-CACRE '16})}. \bibinfo{publisher}{Association for Computing Machinery}, \bibinfo{address}{New York, NY, USA}, Article \bibinfo{articleno}{4}, \bibinfo{numpages}{6}~pages.
\newblock
\showISBNx{9781450342353}
\urldef\tempurl%
\url{https://doi.org/10.1145/2952744.2952748}
\showDOI{\tempurl}


\bibitem[Mohamed and Rezende(2015)]%
        {mohamed_variational_2015}
\bibfield{author}{\bibinfo{person}{Shakir Mohamed} {and} \bibinfo{person}{Danilo~J. Rezende}.} \bibinfo{year}{2015}\natexlab{}.
\newblock \showarticletitle{Variational Information Maximisation for Intrinsically Motivated Reinforcement Learning}. In \bibinfo{booktitle}{\emph{Proceedings of the 28th International Conference on Neural Information Processing Systems - Volume 2}} (Montreal, Canada) \emph{(\bibinfo{series}{NIPS'15})}. \bibinfo{publisher}{MIT Press}, \bibinfo{address}{Cambridge, MA, USA}, \bibinfo{pages}{2125–2133}.
\newblock


\bibitem[Moustris and Tzafestas(2016)]%
        {moustris_intention-based_2016}
\bibfield{author}{\bibinfo{person}{George~P. Moustris} {and} \bibinfo{person}{Costas~S. Tzafestas}.} \bibinfo{year}{2016}\natexlab{}.
\newblock \showarticletitle{Intention-based front-following control for an intelligent robotic rollator in indoor environments}. In \bibinfo{booktitle}{\emph{2016 IEEE Symposium Series on Computational Intelligence (SSCI)}} (Athens, Greece). \bibinfo{publisher}{IEEE}, \bibinfo{address}{New York, NY, USA}, \bibinfo{pages}{1--7}.
\newblock
\urldef\tempurl%
\url{https://doi.org/10.1109/SSCI.2016.7850067}
\showDOI{\tempurl}


\bibitem[Orange and Abaid(2 01)]%
        {orange_transfer_2015}
\bibfield{author}{\bibinfo{person}{N. Orange} {and} \bibinfo{person}{N. Abaid}.} \bibinfo{year}{2015-12-01}\natexlab{}.
\newblock \showarticletitle{A transfer entropy analysis of leader-follower interactions in flying bats}.
\newblock \bibinfo{journal}{\emph{The European Physical Journal Special Topics}} \bibinfo{volume}{224}, \bibinfo{number}{17} (\bibinfo{year}{2015-12-01}), \bibinfo{pages}{3279--3293}.
\newblock
\showISSN{1951-6401}
\urldef\tempurl%
\url{https://doi.org/10.1140/epjst/e2015-50235-9}
\showDOI{\tempurl}


\bibitem[Ortenzi et~al\mbox{.}(1 12)]%
        {ortenzi_object_2021}
\bibfield{author}{\bibinfo{person}{Valerio Ortenzi}, \bibinfo{person}{Akansel Cosgun}, \bibinfo{person}{Tommaso Pardi}, \bibinfo{person}{Wesley~P. Chan}, \bibinfo{person}{Elizabeth Croft}, {and} \bibinfo{person}{Dana Kulić}.} \bibinfo{year}{2021-12}\natexlab{}.
\newblock \showarticletitle{Object Handovers: A Review for Robotics}.
\newblock \bibinfo{journal}{\emph{{IEEE} Transactions on Robotics}} \bibinfo{volume}{37}, \bibinfo{number}{6} (\bibinfo{year}{2021-12}), \bibinfo{pages}{1855--1873}.
\newblock
\showISSN{1941-0468}
\urldef\tempurl%
\url{https://doi.org/10.1109/TRO.2021.3075365}
\showDOI{\tempurl}
\newblock
\shownote{Conference Name: {IEEE} Transactions on Robotics}.


\bibitem[Porfiri(2018)]%
        {porfiri_inferring_2018}
\bibfield{author}{\bibinfo{person}{Maurizio Porfiri}.} \bibinfo{year}{2018}\natexlab{}.
\newblock \showarticletitle{Inferring causal relationships in zebrafish-robot interactions through transfer entropy: a small lure to catch a big fish.}
\newblock \bibinfo{journal}{\emph{{AB}\&C}} \bibinfo{volume}{5}, \bibinfo{number}{4} (\bibinfo{year}{2018}), \bibinfo{pages}{341--367}.
\newblock
\showISSN{23725052, 23724323}
\urldef\tempurl%
\url{https://doi.org/10.26451/abc.05.04.03.2018}
\showDOI{\tempurl}


\bibitem[Rios-Martinez et~al\mbox{.}(2015)]%
        {rios-martinez_proxemics_2015}
\bibfield{author}{\bibinfo{person}{J. Rios-Martinez}, \bibinfo{person}{A. Spalanzani}, {and} \bibinfo{person}{C. Laugier}.} \bibinfo{year}{2015}\natexlab{}.
\newblock \showarticletitle{From Proxemics Theory to Socially-Aware Navigation: A Survey}.
\newblock \bibinfo{journal}{\emph{Int J of Soc Robotics}} \bibinfo{volume}{7}, \bibinfo{number}{2} (\bibinfo{year}{2015}), \bibinfo{pages}{137--153}.
\newblock
\showISSN{1875-4805}
\urldef\tempurl%
\url{https://doi.org/10.1007/s12369-014-0251-1}
\showDOI{\tempurl}


\bibitem[Schreiber(2000)]%
        {schreiber_measuring_2000}
\bibfield{author}{\bibinfo{person}{Thomas Schreiber}.} \bibinfo{year}{2000}\natexlab{}.
\newblock \showarticletitle{Measuring Information Transfer}.
\newblock \bibinfo{journal}{\emph{Phys. Rev. Lett.}} \bibinfo{volume}{85}, \bibinfo{number}{2} (\bibinfo{year}{2000}), \bibinfo{pages}{461--464}.
\newblock
\urldef\tempurl%
\url{https://doi.org/10.1103/PhysRevLett.85.461}
\showDOI{\tempurl}


\bibitem[Shaffer and Abaid(2020)]%
        {shaffer_transfer_2020}
\bibfield{author}{\bibinfo{person}{Irena Shaffer} {and} \bibinfo{person}{Nicole Abaid}.} \bibinfo{year}{2020}\natexlab{}.
\newblock \showarticletitle{Transfer Entropy Analysis of Interactions between Bats Using Position and Echolocation Data}.
\newblock \bibinfo{journal}{\emph{Entropy}} \bibinfo{volume}{22}, \bibinfo{number}{10} (\bibinfo{year}{2020}), \bibinfo{pages}{1176}.
\newblock
\showISSN{1099-4300}
\urldef\tempurl%
\url{https://doi.org/10.3390/e22101176}
\showDOI{\tempurl}


\bibitem[Shannon(1948)]%
        {shannon_mathematical_1948}
\bibfield{author}{\bibinfo{person}{C.~E. Shannon}.} \bibinfo{year}{1948}\natexlab{}.
\newblock \showarticletitle{A Mathematical Theory of Communication}.
\newblock \bibinfo{journal}{\emph{Bell System Technical Jour.}} \bibinfo{volume}{27}, \bibinfo{number}{3} (\bibinfo{year}{1948}), \bibinfo{pages}{379--423}.
\newblock
\showISSN{1538-7305}
\urldef\tempurl%
\url{https://doi.org/10.1002/j.1538-7305.1948.tb01338.x}
\showDOI{\tempurl}


\bibitem[Sumioka et~al\mbox{.}(2007)]%
        {sumioka_causality_2007}
\bibfield{author}{\bibinfo{person}{Hidenobu Sumioka}, \bibinfo{person}{Yuichiro Yoshikawa}, {and} \bibinfo{person}{Minoru Asada}.} \bibinfo{year}{2007}\natexlab{}.
\newblock \showarticletitle{Causality detected by transfer entropy leads acquisition of joint attention}. In \bibinfo{booktitle}{\emph{2007 {IEEE} 6th Int. Conf. on Development and Learning}} (London, UK). \bibinfo{publisher}{IEEE}, \bibinfo{address}{New York, NY, USA}, \bibinfo{pages}{264--269}.
\newblock
\urldef\tempurl%
\url{https://doi.org/10.1109/DEVLRN.2007.4354069}
\showDOI{\tempurl}


\bibitem[Takens(1981)]%
        {takens_detecting_1981}
\bibfield{author}{\bibinfo{person}{Floris Takens}.} \bibinfo{year}{1981}\natexlab{}.
\newblock \showarticletitle{Detecting strange attractors in turbulence}. In \bibinfo{booktitle}{\emph{Dynamical Systems and Turbulence, Warwick 1980}}, \bibfield{editor}{\bibinfo{person}{David Rand} {and} \bibinfo{person}{Lai-Sang Young}} (Eds.). \bibinfo{publisher}{Springer Berlin Heidelberg}, \bibinfo{address}{Berlin, Heidelberg}, \bibinfo{pages}{366--381}.
\newblock
\showISBNx{978-3-540-38945-3}


\bibitem[Thompson et~al\mbox{.}(1992)]%
        {thompson_voluntary_1992}
\bibfield{author}{\bibinfo{person}{P.~D. Thompson}, \bibinfo{person}{J.~G. Colebatch}, \bibinfo{person}{P. Brown}, \bibinfo{person}{J.~C. Rothwell}, \bibinfo{person}{B.~L. Day}, \bibinfo{person}{J.~A. Obeso}, {and} \bibinfo{person}{C.~D. Marsden}.} \bibinfo{year}{1992}\natexlab{}.
\newblock \showarticletitle{Voluntary stimulus-sensitive jerks and jumps mimicking myoclonus or pathological startle syndromes}.
\newblock \bibinfo{journal}{\emph{Movement Disorders}} \bibinfo{volume}{7}, \bibinfo{number}{3} (\bibinfo{year}{1992}), \bibinfo{pages}{257--262}.
\newblock
\showISSN{1531-8257}
\urldef\tempurl%
\url{https://doi.org/10.1002/mds.870070312}
\showDOI{\tempurl}


\bibitem[Tomari et~al\mbox{.}(2014)]%
        {tomari_analysis_2014}
\bibfield{author}{\bibinfo{person}{Razali Tomari}, \bibinfo{person}{Yoshinori Kobayashi}, {and} \bibinfo{person}{Yoshinori Kuno}.} \bibinfo{year}{2014}\natexlab{}.
\newblock \showarticletitle{Analysis of Socially Acceptable Smart Wheelchair Navigation Based on Head Cue Information}.
\newblock \bibinfo{journal}{\emph{Procedia Computer Science}}  \bibinfo{volume}{42} (\bibinfo{year}{2014}), \bibinfo{pages}{198--205}.
\newblock
\showISSN{1877-0509}
\urldef\tempurl%
\url{https://doi.org/10.1016/j.procs.2014.11.052}
\showDOI{\tempurl}


\bibitem[Tranberg~Hansen et~al\mbox{.}(2009)]%
        {tranberg_hansen_adaptive_2009}
\bibfield{author}{\bibinfo{person}{Soren Tranberg~Hansen}, \bibinfo{person}{Mikael Svenstrup}, \bibinfo{person}{Hans~Jorgen Andersen}, {and} \bibinfo{person}{Thomas Bak}.} \bibinfo{year}{2009}\natexlab{}.
\newblock \showarticletitle{Adaptive human aware navigation based on motion pattern analysis}. In \bibinfo{booktitle}{\emph{{RO}-{MAN} 2009}} (Toyama, Japan). \bibinfo{publisher}{IEEE}, \bibinfo{address}{New York, NY, USA}, \bibinfo{pages}{927--932}.
\newblock
\urldef\tempurl%
\url{https://doi.org/10.1109/ROMAN.2009.5326212}
\showDOI{\tempurl}


\bibitem[Urakami and Seaborn(3 15)]%
        {urakami_nonverbal_2023}
\bibfield{author}{\bibinfo{person}{Jacqueline Urakami} {and} \bibinfo{person}{Katie Seaborn}.} \bibinfo{year}{2023-03-15}\natexlab{}.
\newblock \showarticletitle{Nonverbal Cues in Human–Robot Interaction: A Communication Studies Perspective}.
\newblock \bibinfo{journal}{\emph{{ACM} Transactions on Human-Robot Interaction}} \bibinfo{volume}{12}, \bibinfo{number}{2} (\bibinfo{year}{2023-03-15}), \bibinfo{pages}{22:1--22:21}.
\newblock
\urldef\tempurl%
\url{https://doi.org/10.1145/3570169}
\showDOI{\tempurl}


\bibitem[Vinciarelli(2009)]%
        {vinciarelli_social_2008}
\bibfield{author}{\bibinfo{person}{Alessandro Vinciarelli}.} \bibinfo{year}{2009}\natexlab{}.
\newblock \showarticletitle{Social Computers for the Social Animal: State-of-the-Art and Future Perspectives of Social Signal Processing}. In \bibinfo{booktitle}{\emph{User Modeling, Adaptation, and Personalization}}, \bibfield{editor}{\bibinfo{person}{Geert-Jan Houben}, \bibinfo{person}{Gord McCalla}, \bibinfo{person}{Fabio Pianesi}, {and} \bibinfo{person}{Massimo Zancanaro}} (Eds.). \bibinfo{publisher}{Springer Berlin Heidelberg}, \bibinfo{address}{Berlin, Heidelberg}, \bibinfo{pages}{1--1}.
\newblock
\showISBNx{978-3-642-02247-0}
\urldef\tempurl%
\url{https://doi.org/10.1007/978-3-642-02247-0_1}
\showDOI{\tempurl}


\bibitem[Xie et~al\mbox{.}(2022)]%
        {xie_detecting_2022}
\bibfield{author}{\bibinfo{person}{Wei Xie}, \bibinfo{person}{Dongli Gao}, {and} \bibinfo{person}{Eric~Waiming Lee}.} \bibinfo{year}{2022}\natexlab{}.
\newblock \showarticletitle{Detecting Undeclared-Leader-Follower Structure in Pedestrian Evacuation Using Transfer Entropy}.
\newblock \bibinfo{journal}{\emph{{IEEE} Trans. on Intelligent Transportation Systems}} \bibinfo{volume}{23}, \bibinfo{number}{10} (\bibinfo{year}{2022}), \bibinfo{pages}{17644--17653}.
\newblock
\showISSN{1558-0016}
\urldef\tempurl%
\url{https://doi.org/10.1109/TITS.2022.3161813}
\showDOI{\tempurl}


\bibitem[Yang et~al\mbox{.}(2021)]%
        {yang_dataset_2021}
\bibfield{author}{\bibinfo{person}{Fangkai Yang}, \bibinfo{person}{Yuan Gao}, \bibinfo{person}{Ruiyang Ma}, \bibinfo{person}{Sahba Zojaji}, \bibinfo{person}{Ginevra Castellano}, {and} \bibinfo{person}{Christopher Peters}.} \bibinfo{year}{2021}\natexlab{}.
\newblock \showarticletitle{A dataset of human and robot approach behaviors into small free-standing conversational groups}.
\newblock \bibinfo{journal}{\emph{{PLOS} {ONE}}} \bibinfo{volume}{16}, \bibinfo{number}{2} (\bibinfo{year}{2021}), \bibinfo{pages}{e0247364}.
\newblock
\showISSN{1932-6203}
\urldef\tempurl%
\url{https://doi.org/10.1371/journal.pone.0247364}
\showDOI{\tempurl}


\end{thebibliography}

\end{document}